\documentclass[applsci,article,accept,pdftex,moreauthors]{Definitions/mdpi} 

\firstpage{1} 
\pubvolume{14}
\issuenum{15}
\articlenumber{6497}
\pubyear{2024}
\copyrightyear{2024}
\externaleditor{Academic Editor: Andrea Prati}
\datereceived{24 June 2024} 
\daterevised{ 18 July 2024} 
\dateaccepted{24 July 2024 } 
\datepublished{25 July 2024} 
\hreflink{https://doi.org/10.3390/\\ app14156497} 

\Title{Automatic Speech Recognition Advancements for Indigenous Languages of the Americas}

\TitleCitation{Automatic Speech Recognition Advancements for Indigenous Languages of the Americas}

\Author{{Monica} 
 Romero *\orcidA{}, Sandra Gómez-Canaval \orcidB{} and {Ivan G. Torre} 
 \orcidC{}}

\AuthorNames{Monica Romero, Sandra Gómez-Canaval and Ivan G. Torre}

\AuthorCitation{{Romero, M.;} 
 Gómez- Canaval, S.; Torre, I.G.}

 \address[1]{ETS of Computer Systems Engineering, 
 {Universidad} 
 Politécnica de Madrid, 28031 Madrid, Spain; {sm.gomez@upm.es (S.G.-C.)} 
}

\corres{\hangafter=1 \hangindent=1.05em \hspace{-0.82em}Correspondence: monica.romero@alumnos.upm.es}

\abstract{Indigenous languages are a fundamental legacy in the development of human communication, embodying the unique identity and culture of local communities in America. The Second AmericasNLP 
Competition Track 1 of NeurIPS 
2022 proposed the task of training automatic speech recognition (ASR) systems for five Indigenous languages: Quechua, Guarani, Bribri, Kotiria, and Wa'ikhana. In this paper, we describe the fine-tuning of a state-of-the-art ASR model for each target language, using approximately $36.65$ h of transcribed speech data from diverse sources enriched with data augmentation methods. We systematically investigate, using a Bayesian search, the impact of the different hyperparameters on the Wav2vec2.0 XLS-R 
variants of 300 M and 1 B parameters. Our findings indicate that data and detailed hyperparameter tuning significantly affect ASR accuracy, but language complexity determines the final result. The Quechua model achieved the lowest character error rate (CER) ($12.14$), while the Kotiria model, despite having the most extensive dataset during the fine-tuning phase, showed the highest CER ($36.59$). Conversely, with the smallest dataset, the Guarani model achieved a CER of $15.59$, while Bribri and Wa'ikhana obtained, respectively, CERs of $34.70$ and $35.23$. Additionally, Sobol' sensitivity analysis highlighted the crucial roles of freeze fine-tuning updates and dropout rates. We release our best models for each language, marking the first open ASR models for Wa'ikhana and Kotiria. This work opens avenues for future research to advance ASR techniques in preserving minority Indigenous languages.}


\begin{document}

\section{Introduction} 
\label{sec1}

Indigenous languages are natural languages that have linguistically evolved in a particular region attributed to a specific community~\cite{Thiede2020}. They are a paramount heritage of the evolution of language, representing the identity and culture of the local communities, and their cultural contributions constitute an immensely valuable legacy for society~\cite{UNESCO2021}. The lexicon and grammar of Indigenous languages contain knowledge about local ecosystems, traditional techniques, spiritual beliefs, and political organization~\cite{mcquown1955indigenous}. These languages convey a diverse, rich, and ancient narrative characterized by pluralism, heterogeneity, and depth~\cite{UNESCO2021}. They encapsulate the wisdom and accumulated knowledge of generations, often closely tied to the land, natural resources, and the local and ecological constraints~\cite{Chiblow2022}. However, the small number of speaking inhabitants, the absence, in many cases, of writing traditions, the pressure of the dominant languages, or even the dissolution of the native communities have led to a continuous and progressive extinction of Indigenous languages during recent decades~\cite{UNESCO2022}. Actually, $7000$ languages are spoken worldwide, and $6000$ of these are considered Indigenous languages. However, nearly half are endangered, and approximately $1500$ are at extreme risk of extinction~\cite{Bromham2021}. Globalization and technological advances in artificial intelligence (AI) have further accelerated the hazard to minority languages, because solutions based on natural language processing (NLP) and AI are only available for a few dozen languages~\cite{Bromham2021}. Preserving Indigenous languages is an objective that safeguards local communities' heritage and cultural identity and is crucial for retaining critical knowledge associated with environmental interaction and understanding the social aspects of language evolution~\cite{Ferguson2020}.

To address the gap between AI solutions for majority and Indigenous languages, the initiative Second AmericasNLP Competition Track 1 of NeurIPS 2022~\cite{Neurips2022} proposed that the participants produce ASR systems for five different Indigenous languages: Quechua, Guarani, Bribri, Kotiria, and Wa’ikhana. Bribri is a language of the Chibchan family, spoken by approximately $7500$ speakers in areas of Costa Rica and Panama~\cite{Constenla2012, Feldman2020, Kan2022}. Meanwhile, Guarani is spoken in Paraguay, Argentina, Bolivia, and Brazil and is one of the most significant Amerindian languages in terms of numerical importance, with 6--10~million \mbox{speakers~\cite{Adelaar2006, Kan2022, Costa2020}}. Kotiria---or Wanano---is one of the sixteen languages of the eastern branch of the Tukanoan language family~\cite{Stenzel2008}, and it is spoken in the northwest of the Amazon region, in the territory between Brazil and Colombia. Approximately $2000$~speakers are distributed between Colombia and Brazil~\cite{ELP2023, Crevels2012}. Wa'ikhana---or Piratapuyo---is also an eastern Tukanoan language spoken in Brazil and Colombia, with less than \mbox{$2000$ speakers}~\cite{Ethnologue2023, ELP2023, UNESCOP2023}. Finally, Quechua is a heterogeneous language spoken across the Andes, stretching from southern Colombia to northwest Argentina, encompassing a diverse family of languages and dialects~\cite{Pearce2011}, with more than 6.5~million speakers~\cite{Lagos2013}.

Indigenous languages and, more generally, languages with limited or relatively fewer data available are denoted in AI and NLP as low-resource languages with a critical level of under-documentation~\cite{Bromham2021}. They pose a significant challenge for developing robust and accurate ASR systems, as actual algorithms primarily rely on the availability of substantial amounts of labeled data. Despite the difficulty of the task, ASR for low-resource languages has attracted the attention of the scientific community, who have proposed new approaches to this issue, including speech augmentation techniques~\cite{ko2015audio, ko2017study, park2019specaugment}, semi-supervised models~\cite{synnaeve2019end, xu2020iterative, Wav2vec2, Wav2vec2lowresources, parikh-etal-2023-comparing}, fully unsupervised learning methods~\cite{baevski2021unsupervised}, transfer learning techniques~\cite{wang2015transfer, kunze2017transfer, yi2018language}, and specialized neural network architectures~\cite{pmlr-v202-yu23l}, among other solutions.

In this paper, we present our winning approach in the ASR subtask of the America's Challenge competition of NeurIPS 2022~\cite{ebrahimi2022findings}. For this purpose, we trained and optimized an ASR system for each of the following languages: Bribri, Guarani, Kotiria, Wa'ikhana, and Quechua. This marks the first time an ASR model has been developed for the Wa'ikhana and Kotiria languages and we report the first results for these languages in the literature. We addressed the challenge of limited training data by leveraging a semi-supervised model and subsequent fine-tuning using the Wav2vec2.0 framework and applying speed augmentation techniques. The training phase involved meticulous model selection based on the optimization of performance metric hyperparameters. Additionally, we constructed comprehensive n-gram language models using text corpora for the decoding, but the Greedy Search algorithm, supplemented with heuristic corrections, showed better preciseness. Our ASR system showed an average character error rate (CER) of 26.85, thereby achieving the best solution in the competition.

The rest of the paper is organized as follows: Section \ref{sec2} reviews previous research carried out in ASR for Indigenous languages. Section \ref{sec3} details the experimental setup adopted for the architecture, the techniques applied during training, and the dataset creation. Section \ref{sec4} shows the experimental results for our best models with the most suitable hyperparameters and configurations. Finally, in Section \ref{sec5}, we discuss the results and extract the highlights of this research.

\section{Related Work}\label{sec2}

A limited number of studies focus on machine learning techniques applied to the Indigenous languages of Latin America, most of which focus on corpora analysis, NLP applications~\cite{mager2018}, machine translation (MT) systems~\cite{Kan2022,mager2021findings}, sentimental analysis~\cite{aguero2022machine}, or just showing the challenges of this task~\cite{gasser2006machine}. However, only recent initiatives have aimed to address the challenge of developing ASR systems~\cite{jimerson-etal-2023-unhelpful} for the indigenous languages of America. This issue is shared with other minority languages worldwide, and recent initiatives have been carried out to develop ASR models for the Adi (India)~\cite{sasmal2022robust} or for the Cook Islands Māori~\cite{coto2022development} among others. For the languages that concern us in this work, there are no previously reported ASR systems for two of them: Wa'ikhana and Kotiria. In the following paragraph, we summarize the most significant investigations reported in the literature in the area of ASR for Quechua, Bribri, and Guarani. 

The very first ASR system for the Quechua language was published in 2018, and it was able to recognize spoken numbers from one to ten with an accuracy greater than \mbox{$90$\%~\cite{chuctaya2018isolated}}. It was based on cepstral coefficients (MFCC) features, dynamic time warping (DTW), and k-nearest neighbor (KNN) for classifying the characteristics. Then, the first complete E2E 
ASR system in Quechua was created, based on a pre-trained model and fine-tuned in a very limited domain of available data; the results were difficult to extrapolate to other domains~\cite{adams2019massively}. The most detailed ASR system of Quechua to date uses the hidden Markov model toolkit (HTK) and evaluates the improvement when using the Gaussian HMM 
versus the monophonic acoustic model~\cite{zevallos2019automatic}. Later research shows that ASR in Quechua can be improved by adding to the training set's text to speech (TTS) synthetically generated voices~\cite{zevallos2022data}. Similarly to Quechua, there are not many reported investigations on ASR in the Guarani language. The very first reported ASR system was based on Gaussian mixture models (GMM), HMM and n-grams~\cite{maldonado2016ene} and trained and tested with a very limited amount of data. In 2020 and 2021, the OpenASR challenge for low-resource languages included Guarani, increasing the interest in this language~\cite{peterson2021openasr20, peterson2022openasr21}. Previous investigations included architectures based on convolutional neural networks (CNN) and factored time delay neural networks (TDNN-F), which significantly improved the previously reported results~\cite{TDNN}. A more recent work compared three different pre-trained models fine-tuned with 10 h of labeled Guarani speech data~\cite{zhao2022improving}. In 2017, an automatic text-to-voice aligner for the Bribri language, with acoustic models trained on other languages, was published~\cite{Coto2017}. 
Initial explorations into ASR focused on Bribri investigated how vowel--tone separation can improve the accuracy of the ASR models when training data are scarce~\cite{coto-solano-2021-explicit}. Finally, in 2023, the first ASR E2E model based on self-supervised architecture for Bribri was reported during the ASRU2023 ML-SUPERB Challenge 
\cite{chen2023evaluating}.

Integrating ASR with other language processing techniques, such as word embedding, and adopting a linguistic lens could significantly improve outcomes~\cite{coto-solano-2022-evaluating, endangered, Krasnoukhova}, but this took a significant turn with the advent of self-supervised learning models, such as \mbox{Wav2vec2.0~\cite{Wav2vec2, Unsupervised}}. These models capitalized on the power of contrastive learning to transform raw audio data into valuable speech representations, mitigating the need for extensive transcribed data, a common challenge in Indigenous language contexts. The path forward led to exploring Wav2vec2.0's potential for low-resource language scenarios~\cite{Wav2vec2lowresources}, and it demonstrated remarkable adaptability and significant potential in addressing these languages~\cite{MonolingualWav2vec2,interspeech} or other low-resource domains~\cite{torre2021improving, aphasiabenchmark}. More recently, weak supervision emerged as a viable approach for leveraging large-scale, weakly supervised learning to harness the abundant yet imperfectly transcribed public data~\cite{whisper}.

It can be concluded that, although ASR on Indigenous languages has recently received some attention from the scientific community, there are no standardized benchmarks for the languages under study and no reported benchmarks for two of them. In this work, we improve this situation by reporting the best benchmarks for the languages under study in the NeurIPS competition and by sharing with the community the granularity of the many training configurations experimented with, as well as the trained models.


\section{Experimental Setup and Dataset Description}\label{sec3}

\subsection{Main Architecture and Pre-Trained Models}

\begin{figure}[H]
 
    \includegraphics[width=0.9\textwidth]{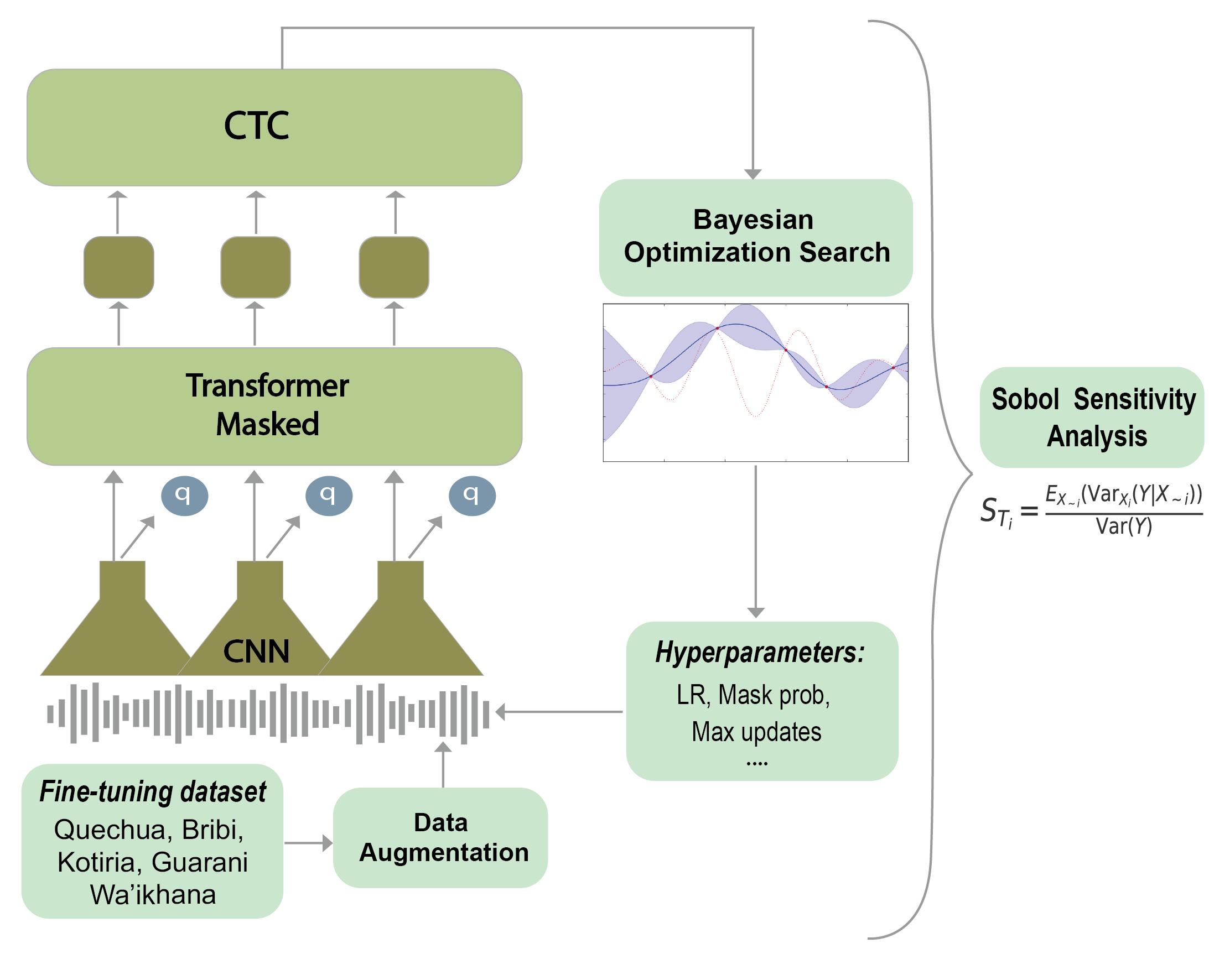}
    \caption{Sketch of the dataset used for fine-tuning the ASR system, the~CNN and transformer-based architecture wav2vec2.0, the~fine-tuning process, the~Bayesian hyperparameter search and the Sobol sensitivity~analysis.}
    \label{fig:wav2vec2-architecture}
\end{figure}

The Wav2Vec 2.0 architecture, illustrated in Figure \ref{fig:wav2vec2-architecture}, constitutes a robust framework for ASR tasks. It comprises three core components: a CNN-based encoder network, a transformer-based context network, and a vector quantization module. These components work in tandem to transform raw audio samples, denoted as $x_i \in X$, into latent speech representations ($z_1, z_2,\ldots, z_T$)~\cite{Wav2vec2}. First, the encoder network, denoted as $f: X \rightarrow Z$, plays a pivotal role in this architecture. It consists of seven sequential blocks of temporal convolution layers, each equipped with 512 channels, strategic strides (5,2,2,2,2,2,2), and kernel sizes (10,3,3,3,3,2,2). This configuration allows the encoder to compress approximately \mbox{25 milliseconds} of 16 kHz audio data into latent representations every 20 milliseconds. Following each convolution layer, layer normalization and Gaussian error linear unit (GELU) activation are applied to enhance feature extraction. The context network, represented as $g: Z \rightarrow C$, takes these latent representations ($z_i,\ldots, z_T$) and builds context representations ($c_i$) that encapsulate the contextual information across the entire sequence of latent speech representations. Here, we leverage two different model sizes: one with \mbox{300~million} (300~M) parameters and the other with 1 billion (1~B) parameters. The context network comprises \mbox{24 blocks} (48 for the bigger model), each with a model dimension of 1024, an inner dimension of 4096, and 16 attention heads. This configuration enables it to capture intricate temporal and contextual dependencies within the audio data~\cite{Wav2vec2}.
\vspace{-4pt}

In the fine-tuning phase of our study, we evaluated two versions of the XLS-R (cross-lingual speech recognition) model framework: XLS-R-300M and XLS-R-1B~\cite{xlsr}. The numbers refer to the quantity of trainable parameters that each model possesses. These models were initially pre-trained on a vast dataset consisting of 436,000 h of publicly available multilingual speech data, in $128$ languages, sourced from various repositories, including CommonVoice~\cite{commonvoice}, Babel~\cite{babel}, Multilingual Librispeech (MLS)~\cite{libspeech}, VoxPopuli~\cite{voxpopuli}, and the VoxLingua107~\cite{voxlingua107} datasets. During the fine-tuning phase, we assessed the performance of the XLS-R-300M and XLS-R-1B models for each Indigenous language separately, utilizing the available training data for each language.

\subsection{Data}
For this work, we have used transcribed speech data collected from the following two~sources: (i) the primary dataset provided by the organization of the AmericasNLP challenge and (ii) a corpus collected from other publicly available sources. The NeurIPS competition advocated using ancillary external resources to complement the primary dataset during the training, but imposed usage restrictions on some databases. A summarized description of the train-dev-test database is depicted in Table \ref{tab1}.

\begin{table}[H]
\caption{Detailed description of train-dev-test splits for each language and data source. {Primary} refers to the corpus provided by the AmericasNLP organization. Speed augmentation is performed only on the primary database and {{external} 
}refers to corpus collected from other sources.}
\label{tab1}
\setlength{\cellWidtha}{\textwidth/7-2\tabcolsep-0in}
\setlength{\cellWidthb}{\textwidth/7-2\tabcolsep-0in}
\setlength{\cellWidthc}{\textwidth/7-2\tabcolsep+0.2in}
\setlength{\cellWidthd}{\textwidth/7-2\tabcolsep-0.1in}
\setlength{\cellWidthe}{\textwidth/7-2\tabcolsep-0.1in}
\setlength{\cellWidthf}{\textwidth/7-2\tabcolsep-0in}
\setlength{\cellWidthg}{\textwidth/7-2\tabcolsep-0in}
\scalebox{1}[1]{\begin{tabularx}{\textwidth}{>{\centering\arraybackslash}m{\cellWidtha}>{\centering\arraybackslash}m{\cellWidthb}>{\centering\arraybackslash}m{\cellWidthc}>{\centering\arraybackslash}m{\cellWidthd}>{\centering\arraybackslash}m{\cellWidthe}>{\centering\arraybackslash}m{\cellWidthf}>{\centering\arraybackslash}m{\cellWidthg}}
\toprule

& \multicolumn{4}{c}{\textbf{Train}} & \textbf{Dev}& \textbf{Test}\\ \cmidrule{2-5}
\textbf{} & \textbf{Primary} & \textbf{Speed Augm.} & \textbf{External} & \textbf{Total} & \textbf{Primary} & \textbf{Primary} \\ \midrule

{{Bribri} 
}& $0.49$ h& $0.98$ h & $1.14$ h& $2.61$ h & $0.04$ h & $0.19$ h\\
{{Guarani} 
}& $0.32$ h& $0.64$ h& - & $0.97$ h & $0.02$ h & $0.12$ h\\
{{Kotiria} 
}& $2.69$ h & $5.43$ h & $21.8$ h & $29.92$ h & $0.5$ h & $0.3$ h \\
{{Wai'khana} 
}& $1.49$ h & $2.98$ h& - & $4.39$ h& $0.1$ h & $0.21$ h\\
{{Quechua} 
}& $1.67$ h & $3.34$ h & $7.04$ h& $12.09$ h& $0.2$ h & $2.08$ h\\ \bottomrule
\end{tabularx}}

\end{table}

\textls[-10]{The database published by the competition included $0.72$ h from The Pandialectal Corpus of the Bribri Language~\cite{bribri}, $0.46$ h from the Guarani Common Voice Mozilla database~\cite{commonvoice}, \mbox{$3.49$ h} from the Endangered Languages Archive (ELAR) (dk0137 and dk0491) for \mbox{Kotiria~\cite{dk0491,dk0137}}}, $1.8$ h from ELAR for Wa'ikhana~\cite{dk0137}, and $3.95$ h from the Siminchik dataset for Quechua~\cite{Siminchikkunarayku}. All of these mentioned databases were subject to restrictions during the competition; it was not permitted to use data from those sources differently than set out by the rules provided by the organization. 
In addition to the primary dataset, we enhanced the training process by collecting supplementary speech data. Specifically, we amassed $1.14$ hours of transcribed speech for the Bribri language~\cite{UCR2021}, $21.8$ h for Kotiria~\cite{Bible}, and $7.04$ h for Quechua~\cite{Brown2020}. However, it should be noted that no external data collection was obtained for Guarani or Wa'ikhana.

To augment the diversity and variability of the assembled speech audios, we applied offline speed augmentation techniques~\cite{ko2015audio} to the primary database. Specifically, we augmented the audios at $\times0.9$ and $\times1.1$ speed rates, effectively widening the spectrum of speech patterns encountered during training. This augmentation technique fortifies ASR models and enhances their adaptability to varied speech styles and tempo changes~\cite{ko2015audio}. Summing up, the cumulative duration of recordings in the training dataset varied across different languages as follows (see Table \ref{tab1}): $2.61$ h for Bribri, less than one hour for Guarani, $29.87$ h for Kotiria, $4.35$ h for Wa'ikhana, and $12.05$ h for Quechua.

\subsection{Decoding Strategy and Language Models}
Different strategies were considered during decoding: greedy decoding and beam search n-gram language model decoding. For the beam search, 3-gram and 4-gram KenLM~
\cite{heafield2011kenlm} models were constructed for each language. For each configuration, two~models were trained depending on the data: one included only training transcription data, and the other included primary transcription data and text data from other corpora. Moreover, we constructed extensive language models for each target language by crawling text corpora spanning diverse sources like speech transcriptions, online texts, and books. The size of the text corpora collected for training the language models was relatively modest, with fewer than $100$k words secured for each language. Beam search hyperparameters $\alpha$ and $\beta$ were selected based on the dev set and searched during $50$ trials with Bayesian optimization techniques, by fixing a beam width of $128$. 

However, due to the lack of a standard normalization of the transcriptions and the low amount of data, this optimization did not lead to significant improvement and even degraded performance for some languages. Therefore, the final decoding strategy was only based on greedy search and heuristic corrections applied to correct textual errors such as capitalization, punctuation, and reducing multiple spaces or letters.

\subsection{Hyperparameter Fine-Tuning}

We delved deeper into the experimental findings to understand the contribution of different hyperparameters on the performance of the language models, specifically focusing on the variants of the Wav2vec2.0 XLS-R model: 300 M and 1 B parameters. We studied and analyzed the impact of the most important hyperparameters: learning rate, maximum number of updates, freeze fine-tune updates, activation dropout, mask probability, and mask channel probability. Hyperparameters serve as control parameters that govern various facets of the training process, thereby exerting a profound influence on model performance. However, while the best hyperparameters are usually published, it is not usual to find an exhaustive study dissecting their individual contribution impacts and their complex correlations.

Our methodology for hyperparameter optimization revolves around an exhaustive Bayesian search~\cite{shahriari2015taking} with the help of the Optuna hyperparameter optimization framework~\cite{optuna_2019}. We traverse a wide spectrum of hyperparameter configurations (see Table \ref{tabhyper}), assessing model performance at each juncture in order to determine the most effective parameter settings for our specific task and model variations. Finally, two model checkpoints are considered during the test phase: the one with the lowest loss and the one with the lowest word error rate (WER) during training.

\begin{table}[H]
\caption{Hyperparameter search range during the fine-tuning training phase.}
\label{tabhyper}
\setlength{\cellWidtha}{\textwidth/3-2\tabcolsep-0in}
\setlength{\cellWidthb}{\textwidth/3-2\tabcolsep-0in}
\setlength{\cellWidthc}{\textwidth/3-2\tabcolsep-0in}
\scalebox{1}[1]{\begin{tabularx}{\textwidth}{>{\centering\arraybackslash}m{\cellWidtha}>{\centering\arraybackslash}m{\cellWidthb}>{\centering\arraybackslash}m{\cellWidthc}}
\toprule
& \textbf{Minimum Value} & \textbf{Maximum Value} \\ \midrule
{{Learning rate} 
} & $10^{-6}$ & $10^{-3}$ \\ 
{{Max number of updates} 
} & $10$k & $100$k \\
{{Freeze fine-tune updates} 
}& $0$ & $50$k \\
{{Activation dropout} 
} & $0.01$ & $0.2$ \\
{{Mask probability} 
} & $0.2$ & $0.7$ \\ 
{{Mask channel probability} 
} & $0.1$ & $0.7$ \\ \bottomrule
\end{tabularx}}
\end{table}

\subsection{Sobol' Sensitivity Analysis for Hyperparameter Explanation}

Sobol' sensitivity analysis is a relatively recent technique used for estimating the influence of individual variables on the output of complex mathematical models~\cite{sobol2001global}. While it has been widely adopted across diverse scientific disciplines ranging from epidemiology~\cite{sobol2001global} to ecology~\cite{langie2022toward,schneider2020impact}, 
it has only been recently introduced to the study of AI complex models~\cite{linardatos2020explainable, antoniadis2021random}. We have utilized Sobol' sensitivity analysis to assess the impact of various hyperparameters on the performance of our language models designed for Quechua, Kotiria, Bribri, Guarani, and Wa'ikhana, producing new valuable and fresh insights in the realm of ASR.

Sobol' sensitivity analysis measures how much of the variance in the model output can be attributed to the specific inputs or set of inputs. This approach considers only the inputs and outputs of the system under study, considering the function transformation as a black box. It provides two essential indices: the first-order (S1) index, which quantifies the individual contribution of each input parameter to the output variance, and the total-order (ST) index, which takes into account not only each parameter but also its interactions with all other parameters in the model~\cite{sobol1, sobol2001global}.

\section{Results}\label{sec4}

In this section, we delve into the outcomes of our investigation, where we explore the intricate interplay between hyperparameter configurations and the performance of diverse language models. The ensuing analysis, encapsulated in Table \ref{tab:results}, summarizes the best hyperparameter configuration obtained for each language.

\begin{table}[H]
\caption{WERs and CERs for the best fine-tuning hyperparameter configurations for Quechua, Kotiria, Bribri, Guarani, and Wa'ikhana. Additional configurations and results can be found in the Supplementary Information.}
\label{tab:results}

\begin{adjustwidth}{-\extralength}{0cm}
\setlength{\cellWidtha}{\fulllength/8-2\tabcolsep-0in}
\setlength{\cellWidthb}{\fulllength/8-2\tabcolsep-0in}
\setlength{\cellWidthc}{\fulllength/8-2\tabcolsep-0in}
\setlength{\cellWidthd}{\fulllength/8-2\tabcolsep-0in}
\setlength{\cellWidthe}{\fulllength/8-2\tabcolsep-0in}
\setlength{\cellWidthf}{\fulllength/8-2\tabcolsep-0in}
\setlength{\cellWidthg}{\fulllength/8-2\tabcolsep-0in}
\setlength{\cellWidthh}{\fulllength/8-2\tabcolsep-0in}
\scalebox{1}[1]{\begin{tabularx}{\fulllength}{>{\centering\arraybackslash}m{\cellWidtha}>{\centering\arraybackslash}m{\cellWidthb}>{\centering\arraybackslash}m{\cellWidthc}>{\centering\arraybackslash}m{\cellWidthd}>{\centering\arraybackslash}m{\cellWidthe}>{\centering\arraybackslash}m{\cellWidthf}>{\centering\arraybackslash}m{\cellWidthg}>{\centering\arraybackslash}m{\cellWidthh}}
\toprule

\textbf{Language} & \textbf{Learning Rate} & \textbf{Max Updates} & {\textbf{Freeze Fine-Tune}} & \textbf{Mask Channel} & \textbf{Activation Dropout} & \textbf{WER} & \textbf{CER} \\ 
\midrule
Quechua & $1 \times 10^{-5}$ & $90$k & $5$k & $0.5$ & $0.1$ & $48.98$ & $12.14$\\
Kotiria & $1 \times 10^{-5}$ & $40$k & $5$k & $0.5$ & $0.1$ & $79.69$ & $36.59$ \\ 
Bribri & $1 \times 10^{-4}$ & $90$k & $8$k & $0.3$ & $0.2$ & $69.03$ & $34.70$ \\ 
Guarani & $1 \times 10^{-5}$ & $90$k & $9$k & $0.3$ & $0.1$ & $62.91$ & $15.59$ \\ 
Wa'ikhana & $1 \times 10^{-5}$ & $130$k & $1$k & $0.25$ & $0.1$ & $68.42$ & $35.23$ \\ 
\bottomrule
\end{tabularx}}

\end{adjustwidth}
\end{table}

The Quechua language model, benefiting from a substantial training dataset of \mbox{$12.09$ h}, achieved the best performance, with a WER of $48.98\%$ and a CER of $12.14\%$. This performance could be influenced not only by the amount of data but also by the quality and diversity of the data. However, our investigation also unveiled a counter-intuitive observation: the Kotiria language model, with the most extensive dataset of $29.92$ h (see Table \ref{tab1}), showed the highest WER ($79.69\%$) and CER ($36.59\%$) rates. This discrepancy could arise from language complexity, data quality, or diversity. Remarkably, the Guarani language model, trained on the smallest dataset ($0.97$ h), achieved a WER of $62.91\%$ and a CER of $15.59\%$. The notable effectiveness of meticulous hyperparameter tuning, even with limited data, highlights the importance of this process. The precise adaptation of hyperparameters can somewhat compensate for data scarcity, optimizing model performance. This trend extended to the Bribri and Wa'ikhana language models, which exhibited intermediate performance levels despite varying dataset sizes ($2.38$ and $6.11$ h, respectively). These outcomes, as shown in Table \ref{tab:results}, underscore the pivotal role of hyperparameter tuning in determining model performance. Additional configurations and results can be found in the Supplementary Information.

In addition, our research examined two variants of the Wav2vec2.0 XLS-R model: one with 300~million parameters and another with 1 billion parameters. Despite the difference in parameter count, both models yielded comparable results, except for Kotiria, where the bigger model showed significantly better performance. This suggests that the choice between these two models may be influenced by resource constraints, where the smaller model may be preferable for applications with limited resources, while the larger model could offer slightly improved performance when more data are available.

\begin{figure}[H]
   
    \includegraphics[width=0.9\textwidth]{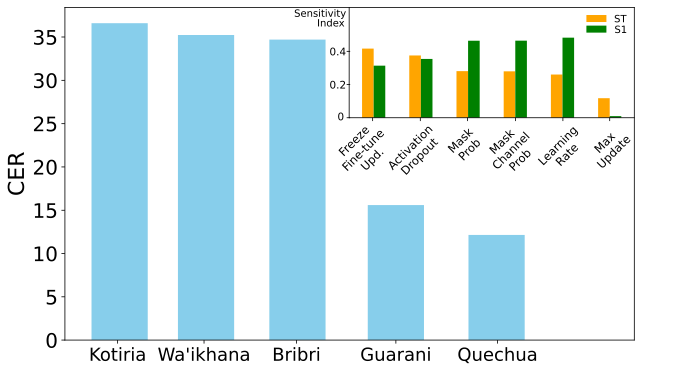}
    \caption{The outer bar chart panel displays the character error rates (CERs) for five Indigenous language models: Kotiria, Wa'ikhana, Bribri, Guarani, and Quechua. Lower bars indicate better-quality performance of the model. The inner panel provides a Sobol' sensitivity analysis of the various hyperparameters tuned during model training, assessing their impact on model performance variability. The orange bars represent the total sensitivity (ST) index, while the green bars indicate the first-order sensitivity (S1) index. A higher bar indicates the more importance of that hyperparameter when correctly choosing it during the fine-tuning phase.}
    \label{fig:wer-sobol}
\end{figure}

\textls[-15]{Analyzing the hyperparameters across different languages, it is noticeable that the learning rate is consistently set at either \(1 \times 10^{-5}\) or \(1 \times 10^{-4}\), although the search range is wider (see Table \ref{tabhyper}). This low learning rate aligns with previously reported \mbox{investigations~\cite{Wav2vec2, baevski2021unsupervised, aphasiabenchmark}}}, allowing precise weight adjustments during the fine-tuning phase. The learning rate is highly related to the maximum number of updates. In this case, the total number of iterations the model trains varies widely, from 40k to 130k. However, no apparent simple relationship exists between the number of updates, the optimization step size, or the available training datasets illustrated in Figure \ref{fig:wer-sobol}. This suggests the requirement of different amounts of training depending upon the complexity and uniqueness of the languages. The fine-tuning freezing point, denoting the end at which specific model layers are frozen and not updated, seems to increase with the maximum number of updates for the Quechua, Bribri, and Guarani languages. However, this pattern does not hold for the Wa'ikhana language, which combines a lower freezing point with more updates. The mask channel values hover between $0.25$ and $0.5$, implying the significance of masking specific inputs during training, potentially reducing overfitting. Finally, activation dropout rates are consistently low across the models, between $0.1$ and $0.2$, suggesting a regularization strategy to prevent overfitting while maintaining the capability to learn complex patterns.

Specific hyperparameters did not correlate directly with lower WER or loss values. This suggests that the optimal hyperparameters are highly dependent on language-specific characteristics. Nonetheless, it is worth noting that, despite similar learning rates, mask channel probabilities, and activation dropout rates, the Guarani language model achieved notably lower WER and loss values, hinting at the influence of factors like dataset quality, model architecture, or inherent language characteristics.

The Sobol' sensitivity analysis reveals that the number of fine-tuning updates and the activation dropout rate are critical for model performance. This indicates that careful attention to these hyperparameters can significantly improve results. Interestingly, the low impact of the maximum number of updates suggests that other factors, such as layer freezing and regularization, are more influential in data-limited contexts.

Table \ref{tab:sensitivity_analysis} presents the outcomes of the Sobol' sensitivity analysis, offering insights into the relative significance of each hyperparameter in influencing the WER of the language model. This analysis delves into the influence of six key hyperparameters: learning rate, max number of updates, freeze fine-tune updates, mask prob, mask channel prob, and activation dropout, on the model's WER. The first-order sensitivity (S1) index measures the individual contribution of the parameter to the total output variance, while the total sensitivity (ST) index reflects the total contribution of a parameter, including its interactions with the other hyperparameters, to the final output. Interestingly, hyperparameters with the most substantial individual influence on the WER (S1), e.g., learning rate (ST = 0.57) or masked channel probability (S1 = 0.54), do not have so high an impact when considering their interactions with other hyperparameters (ST = 0.26 and ST = 0.28, respectively). Notably, the hyperparameter with a higher ST is the freeze fine-tuning updates, suggesting that a bad choice could lead to \textit{{forgetting} 
}the info in the pre-trained layers or not \textit{{adapting} 
}the pre-trained neural network to the current domain. On the other hand, unexpectedly, the maximum number of updates during training seems not so sensible, although it is usually a parameter to which the research community shows more attention. Finally, activation dropout is the second most important contribution to the output variance on the ST, which is somehow expected, as it is a sensible tuning that avoids overfitting but which, in turn, can make training difficult when not properly adjusted.

\begin{table}[H]
\caption{Sensitivity analysis results using first-order (S1) Sobol' and global (ST) indices for the different fine-tuning hyperparameters. Parameters have been ordered by their ST importance.}
\label{tab:sensitivity_analysis}
\setlength{\cellWidtha}{\textwidth/3-2\tabcolsep-0in}
\setlength{\cellWidthb}{\textwidth/3-2\tabcolsep-0in}
\setlength{\cellWidthc}{\textwidth/3-2\tabcolsep-0in}
\scalebox{1}[1]{\begin{tabularx}{\textwidth}{>{\centering\arraybackslash}m{\cellWidtha}>{\centering\arraybackslash}m{\cellWidthb}>{\centering\arraybackslash}m{\cellWidthc}}
\toprule
\textbf{Parameter} & \textbf{S1} & \textbf{ST} \\
\midrule
Freeze fine-tune updates & $0.32$ & $0.41$ \\
Activation dropout & $0.35$ & $0.37$ \\
Mask prob & $0.54$ & $0.29$ \\
Mask channel prob & $0.54$ & $0.28$\\
Learning rate (lr) & $0.57$ & $0.26$ \\
Max update & $\sim$0 & $0.13$ \\
\bottomrule
\end{tabularx}}
\end{table}

\section{Conclusions and Future Work}\label{sec5}
In this work, we have fine-tuned and reported the very first ASR models for the Wa'ikhana and Kotiria languages and have additionally established new benchmarks within the dataset of the Americas NLP Challenge 2022 for Guarani, Quechua, and Bribri, marking a significant step forward in this field. These results represent an important contribution to the study of ASR in Indigenous languages, in which only a few dozen studies are reported for these languages. Additionally, we have published our best models in a repository and the full hyperparameter experiments in the SI. 

We show that pre-trained models are very promising when fine-tuning them to some unknown domain with a very low amount of data but that bigger architectures do not always achieve better results. This may be due to the fact that, when the amount of data is very small and also very far from the pre-trained domain, it is more difficult to converge to an optimal solution if the number of trainable parameters is too large. This would imply that for some corner cases and minority languages, the race to use architectures with an increasing number of parameters may not be the most effective. Surprisingly, we did not find a clear relationship between the number of hours available for each language and the accuracy of the fine-tuned models. 
This may be related to the phonemic distance between the pre-trained languages and the fine-tuned domains, which may have made the quantized latent pre-trained representations unsuitable for all the target languages. 

The application of Sobol' sensitivity analysis has allowed us to identify critical hyperparameters that impact language model performance in ASR tasks. These insights inform future hyperparameter tuning efforts, enhancing the accuracy and efficiency of ASR systems for under-resourced languages. We unveil that, when considering internal correlations, the number of epochs where the pre-trained layers of the models are frozen is significantly more important than the total number of trained epochs. This indicates that, when a low amount of data is available for fine-tuning, the greatest effort has to be made in inefficiently adjusting the weights of the neural network in the last layers rather than in adjusting its weights across the architecture. The second most important factor unveiled by Sobol' analysis is the dropout rate, which shows that, when the number of trainable parameters is so large and the amount of data is so scarce, the risk of overtraining is very high. This comprehensive evaluation of hyperparameters provides a valuable tool for optimizing ASR models, which is particularly crucial for languages with limited resources and data availability. 

While our proposed ASR system demonstrates significant advancements for the Indigenous languages of the Americas, several limitations remain. First, the relatively small size of the training datasets limits the model's ability to generalize across diverse speech patterns and dialects. Additionally, the quality and variety of the data used for training and evaluation can impact the performance, as some languages had fewer resources available. Another limitation is the dependency on pre-trained models, which may not fully capture the unique phonetic and grammatical structures of these languages. Future work should focus on expanding the datasets with more diverse and representative samples, improving data collection methodologies, using transfer learning and knowledge distillation, and exploring different model architectures---such as others based on conformers---that can better handle the linguistic complexities of Indigenous languages~\cite{ieee2024asr}. Additionally, bimodal or multimodal recognition could be particularly beneficial for low-resource language ASR systems~\cite{ieee2023emotion}. Additionally, techniques successfully used in studies focusing on ASR for individuals with speech disorders could also benefit the domain of Indigenous languages, wherein the many of the phonemes are unseen during the pertaining phase and the data size is small~\cite{ieee2021dysarthric}.
Furthermore, incorporating community feedback and collaboration will be essential to ensure these ASR systems are accurate, culturally appropriate, and beneficial for the preservation and revitalization of these languages.

To round off, we hope that this work paves the way for new research avenues in ASR for Indigenous languages and minority domains in general. Future research could explore advanced methods, such as automated machine learning (AutoML), for a more exhaustive exploration of hyperparameter spaces, potentially resulting in more optimized models. Furthermore, advanced regularization techniques could mitigate model complexity and overfitting, especially in languages with limited data. Collecting more data for under-resourced languages is important, but the emphasis should be on capturing linguistic intricacies, dialectal variations, and cultural contexts to train more robust and culturally sensitive models. Furthermore, exploring the potential benefits of transfer learning from high-resource languages to less-resourced languages is an avenue worth pursuing. Cross-linguistic knowledge transfer could help mitigate the challenges posed by data scarcity and enhance model performance for underrepresented languages. These and other possible tasks will help to democratize access to AI speech models for the thousands of languages across the Earth.

\section{Data and Models Accessibility}\label{sec6}
The data utilized for fine-tuning, provided by the AmericasNLP challenge, are accessible online~\cite{romero2024asr}. The results obtained by performing experiments with different hyperparameters and configurations are available in the Supplementary Information (SI). The fine-tuned models for each language, including Quechua, Bribri, Kotiria, Guarani, and Wa'ikhana, are accessible online~\cite{romero2024asr}. The models can be accessed at the following URLs: 
\begin{itemize}
    \item  Quechua (\url{https://huggingface.co/ivangtorre/wav2vec2-xlsr-300m-quechua})   {(accessed on 12 September 2023)} 
    \item Kotiria (\url{https://huggingface.co/ivangtorre/wav2vec2-xlsr-300m-kotiria}) {(accessed on 12 September 2023)} 
    \item Wa'ikhana (\url{https://huggingface.co/ivangtorre/wav2vec2-xlsr-300m-waikhana}) {(accessed on  12 September 2023)} 
    \item Guarani (\url{https://huggingface.co/ivangtorre/wav2vec2-xlsr-300m-guarani}) {(accessed on  12 September 2023)} 
    \item Bribri (\url{https://huggingface.co/ivangtorre/wav2vec2-xlsr-300m-bribri}) {(accessed on  12 September 2023).} 
\end{itemize}

Scripts prepared for downloading the fine-tuned models and performing inference are now available at \url{https://github.com/monirome/asr-indigenous-languages} {(accessed on  12 September 2023)} 
\cite{romero2024asr}.

\vspace{+6pt} 
\supplementary{The following supporting information can be downloaded at: \url{www.mdpi.com/xxx/s1}.  Table S1: Detailed fine-tuning results for the Quechua ASR model, illustrating WER variations based on hyperparameter
configuration. Table S2. Detailed fine-tuning results for the Kotiria ASR model, illustrating WER variations based on hyperparameter
configuration. Table S3. Detailed fine-tuning results for the Bribri ASR model, illustrating WER variations based on hyperparameter
configuration. Table S4. Detailed fine-tuning results for the Guarani ASR model, illustrating WER variations based on hyperparameter
configuration. Table S5. Detailed fine-tuning results for the Wai’khana ASR model, illustrating WER variations based on
hyperparameter~configuration.}

\authorcontributions{Conceptualization, I.G.T.; Methodology, I.G.T.; Software, M.R.; Investigation, M.R. and I.G.T.; Resources, M.R.; Data curation, M.R.; Writing---original draft, M.R. and  I.G.T.; Writing---review \& editing, M.R., S.G.-C. and  I.G.T.; Supervision, S.G.-C. and  I.G.T.; Funding acquisition, S.G.-C.  All authors have read and agreed to the published version of the manuscript.
}

\funding{This research received no external funding. 
}

\institutionalreview{Not applicable. 
}

\informedconsent{Not applicable. 
}

\dataavailability{The original contributions presented in the study are included in the article/Supplementary Materials, further inquiries can be directed to the corresponding authors.

} 

\acknowledgments{The authors gratefully acknowledge the Universidad Politécnica de Madrid for providing computing resources on the Magerit Supercomputer.}

\conflictsofinterest{The authors declare no conflicts of interest.
}

\begin{adjustwidth}{-\extralength}{0cm}
\reftitle{References}

\PublishersNote{}
\end{adjustwidth}

\end{document}